\documentclass[pdflatex,sn-mathphys-num]{sn-jnl}

\usepackage{graphicx}%
\usepackage{orcidlink}
\usepackage{multirow}%
\usepackage{amsmath,amssymb,amsfonts}%
\usepackage{amsthm}%
\usepackage[mathscr]{euscript}
\let\mathscr\EuScript
\usepackage[title]{appendix}%
\usepackage{xcolor}%
\usepackage{textcomp}%
\usepackage{manyfoot}%
\usepackage{booktabs}%
\usepackage{algorithm}%
\usepackage{algorithmicx}%
\usepackage{algpseudocode}%
\usepackage{listings}%
\usepackage{float} 
\usepackage{placeins}
\usepackage[utf8]{inputenc}
\usepackage{textgreek}
\usepackage{hyperref}
\usepackage{graphicx}
\usepackage{amsmath, amssymb}
\usepackage{booktabs}
\usepackage{multirow}
\usepackage{xcolor}
\usepackage{orcidlink}
\usepackage{academicons}

\begin{document}














\title[Explainable AI for Pericoronitis Assessment]{An Explainable Two-Stage Deep Learning Framework for Pericoronitis Assessment in Panoramic Radiographs Using YOLOv8 and ResNet-50}


\author*[1]{\fnm{Ajo} \sur{Babu George} \orcidlink{0009-0005-3026-0959}}\email{drajo\_george@dicemed.com}

\author[2]{\fnm{Pranav} \sur{S} \orcidlink{0009-0001-7611-8596}}\email{pranavsudheesh34@gmail.com}
\equalcont{These authors contributed equally to this work.}

\author[3]{\fnm{Kunal} \sur{Agarwal} \orcidlink{0000-0002-5130-208X}}\email{academicagarwal@gmail.com}
\equalcont{These authors contributed equally to this work.}

\affil*[1]{ \orgname{DiceMed}, \city{Cuttack}, \state{Odisha}, \country{India}}

\affil[2]{\orgdiv{Department of Computer Science and Engineering}, \orgname{College of Engineering Trivandrum},\city{Thiruvananthapuram}, \state{Kerala}, \country{India}}

\affil[3]{\orgdiv{Department of Oral Medicine and Radiology}, \orgname{S.C.B Dental College and Hospital},\city{Cuttack}, \state{Odisha}, \country{India}}


\abstract{\textbf{Objectives:} To overcome challenges in diagnosing pericoronitis on panoramic radiographs (OPGs), an AI-assisted assessment system integrating anatomical localization, pathological classification, and interpretability.

\textbf{Methods:} A two-stage deep learning pipeline was implemented. The first stage used YOLOv8 to detect third molars and classify their anatomical positions and angulations based on Winter’s classification. Detected regions were then fed into a second-stage classifier, a modified ResNet-50 architecture, for detecting radiographic features suggestive of pericoronitis. To enhance clinical trust, Grad-CAM was used to highlight key diagnostic regions on the radiographs.

\textbf{Results:} The YOLOv8 component achieved 92\% precision and 92.5\% mean average precision. The ResNet-50 classifier yielded F1-scores of 88\% for normal cases and 86\% for pericoronitis. Radiologists reported 84\% alignment between Grad-CAM and their diagnostic impressions, supporting the radiographic relevance of the interpretability output.

\textbf{Conclusion:} The system shows strong potential for AI-assisted panoramic assessment, with explainable AI features that support clinical confidence.

\textbf{Clinical Relevance:} An AI system utilizing YOLOv8 and ResNet-50 identifies radiographic patterns associated with pericoronitis, providing anatomical localization, pathological classification, and interpretable heatmaps for enhanced  radiographic assessment.}




\maketitle

\section{Introduction}\label{sec1}

Pericoronitis is an acute inflammatory condition affecting the soft tissues surrounding the crown portions  of a partially erupted tooth, most commonly the third molars. If not diagnosed and treated promptly, pericoronitis can cause severe pain, trismus, abscess formation, and, in rare cases, may progress to life-threatening deep space infections \cite{kourehpaz2024comparison}.

The diagnosis of pericoronitis typically relies on a combination of clinical evaluation and radiographic assessment, particularly intraoral periapical radiographs (IOPAR) and panoramic radiographs (orthopantomograms, OPG). Radiographic features such as position and angulation of the partially erupted third molar, the presence of an operculum, and pericoronal semilunar or circumferential radiolucency  are critical for accurate diagnosis and treatment planning\cite{farah2002pericoronal}. Especially in rural or low-resource settings, where access to experienced oral radiologists is often limited, interpreting these features can be challenging and prone to inter-observer variability, leading to inconsistent diagnoses and delayed care. Delayed diagnosis increases complications and costs. AI-based tools could assist general dentists in low-resource settings by providing consistent, early detection from panoramic radiographs, potentially reducing misdiagnosis and improving treatment outcomes.

Recent advances in artificial intelligence (AI) and deep learning have shown strong potential in automating medical image analysis, including dental radiographs, by detecting abnormalities with speed and consistency. This involves diverse areas such as caries detection, periodontal disease assessment, and tooth segmentation~\citep{Beser2024,Ali2025}. Nevertheless, automated detection of pericoronitis remains a challenging task due to the complex anatomical variations, overlapping structures in panoramic images, and the subtlety of pathological changes. Many existing approaches focus either on tooth detection or disease classification in isolation, without integrating both anatomical localization and pathological assessment within a unified framework~\citep{Sindi2023}.

To address these challenges, a two-stage deep learning pipeline is introduced for the automated detection of pericoronitis in panoramic radiographs. The first stage utilizes the YOLOv8~\citep{widayani2024review} object detection model to localize third molars and extract key anatomical features such as position (quadrant) and angulation (Winter's classification). The second stage applies a ResNet-50~\citep{Wahyuningsih2023} convolutional neural network to classify each detected third molar as healthy or affected by pericoronitis, based on cropped ROI from the original radiograph. To improve transparency and build clinical trust in AI-assisted diagnosis, visual explainability is incorporated using Gradient-weighted Class Activation Mapping (Grad-CAM) to generate visual heatmaps that highlight the regions that are most influential in the model's decision-making process.
By addressing this unmet clinical need, the proposed framework aims to bridge the gap between expert-level radiographic interpretation and everyday dental practice, especially in low resource settings.

The contributions of this work are as follows:
\begin{itemize}
\item Development of an end-to-end pipeline combining anatomical localization and disease classification for pericoronitis detection in panoramic radiographs.
\item Demonstration of the pipeline's effectiveness on a clinically annotated dataset, showing high diagnostic accuracy and strong agreement with expert annotations.
\item Integration of explainable AI techniques to provide visual insight into model predictions, enhancing transparency and clinical usability.
\end{itemize}

\section{Background Study}\label{sec:ai_dentistry}

Table~\ref{tab:ai_dental_methods_tasks} shows the background study performed across 10 related works on AI-based diagnosis using dental panoramic radiographs. Each study is analyzed based on methodology, dataset, and limitations.

\begin{sidewaystable}[htbp]
\caption{Comparison of methodologies, tasks, and limitations in recent maxillofacial radiology AI studies (2021--2025)}
\label{tab:ai_dental_methods_tasks}
\begin{tabular*}{\textheight}{@{\extracolsep\fill} p{6cm} p{5cm} p{6cm}}
\toprule
\textbf{Methodology} & \textbf{Tasks} & \textbf{Limitations} \\
\midrule

MobileNetV2 binary classifier with Grad-CAM \cite{vinayahalingam2021classification} & 
Third molar caries detection (binary classification); interpretability (Grad-CAM) & 
Only third molars; cropped images limit clinical workflow applicability \\

ResNet50 + attention loss with Grad-CAM \cite{gwak2024attention} & 
Jaw lesion classification with minimal annotation; interpretability (Grad-CAM) & 
Manual attention labeling; single institution; limited lesion types \\

Two-branch CNN (classification + segmentation) with Grad-CAM \cite{yu2022deep}& 
General lesion detection and segmentation; interpretability (Grad-CAM) & 
No symptom-level reasoning; lacks external validation \\

ResNet-50 binary caries classifier with Grad-CAM \cite{oztekin2023explainable}& 
Tooth-level caries detection (binary classification); explainable AI (Grad-CAM) & 
Manually cropped teeth; small early-stage caries sample; limited subject pool \\

U²-Net with deep supervision \cite{boztuna2024segmentation} & 
Periapical lesion segmentation & 
Small dataset; lacks CBCT confirmation; single-center \\

YOLOv8 for CEJ/alveolar crest detection \cite{jundaeng2025artificial}& 
Alveolar bone loss estimation; periodontitis staging & 
Moderate expert agreement; lacks early-stage detection sensitivity \\

DenseNet121 for Stafne’s cavity/cyst/tumor differentiation \cite{lee2021deep} & 
Differentiation of Stafne’s bone cavity vs odontogenic cysts/tumors & 
Misclassified rare SBC types; single-center; lacks external testing \\

ResNet18 text classifier with Grad-CAM \cite{zhang2023grad} & 
Text-based dental diagnosis from clinical notes; explainability (Grad-CAM) & 
Not image-based; not dental-specific; lacks radiological context \\
\bottomrule
\end{tabular*}
\end{sidewaystable}

\section{Methods and Methodology}\label{sec2}

\subsection{Dataset Description}

In this retrospective study, a two-stage deep learning pipeline, comprising YOLOv8, ResNet-50, and Grad-CAM, was developed for the automated detection of pericoronitis in panoramic radiographs. The methodology and reporting adhered to the Checklist for Artificial Intelligence in Medical Imaging (CLAIM) and the Standards for the Reporting of Diagnostic Accuracy Studies (STARD) Checklist, ensuring transparency, reproducibility, and clinical relevance. Diagnostic fidelity was maintained by following established radiographic norms for data acquisition, image reconstruction, and cone beam artifact suppression.The curated  dataset, comprising 775 radiographically suggestive  cases of pericoronitis and 775 control samples, provided a robust basis for evaluating model performance with narrow confidence intervals and minimizing overfitting, significantly exceeding the typical recommendation of 100--150 samples per class needed for 95\% confidence with a ±0.05 margin of error in diagnostic AI classification studies \cite{tang2018data, bujang2021step}. Future work proposes integrating an unsupervised active guardrail mechanism to enhance anomaly detection and model reliability, drawing inspiration from similar safety guardrails used in large language models \cite{arun2025integrated}.

\subsection{Dataset Composition}

\subsubsection{Model 1: Localisation Model}
A multicenter cohort of 1,190 de-identified panoramic radiographs was curated from four public repositories (Table~\ref{table}). Images were selected based on strict inclusion criteria, including visible third molars, absence of severe artifacts, and confirmed anatomical coverage from condyle to contralateral premolar. An 80:20 stratified split preserved class balance across quadrants and Winter's classifications.

\subsubsection{Model 2: Classification Model}

From localized ROIs, 1,550 cropped third molar regions (775 pericoronitis / 775 controls) were extracted. To minimize diagnostic ambiguity, only cases with radiographic indicative features of pericoronitis were included. Labeling was independently performed and reconciled by two experienced oral radiologists. Inclusion criteria required signs such as widened follicular space ($>2\text{mm}$), osteolytic changes adjacent to the distal root, and pericoronal radiolucency without sclerotic remodeling. Cases with uncertain chronicity or without radiographic evidence of active inflammation were excluded.

\begin{table}[ht]
\centering
\caption{List of Dental X-ray Datasets}
\label{table}
\begin{tabular}{|l|l|}
\hline
\textbf{Dataset Name} & \textbf{Link} \\
\hline
Dental OPG X-ray Dataset & \href{https://www.kaggle.com/datasets/imtkaggleteam/dental-opg-xray-dataset}{Kaggle} \\
Panoramic Dental X-ray Images & \href{https://www.kaggle.com/datasets/amitabhabu/panoramic-dental-x-ray-images}{Kaggle} \\
Dental X-ray Images & \href{https://data.mendeley.com/datasets/xn5bz6fdm6/1}{Mendeley Data} \\
STS-Tooth: A Multi-Modal Dental Dataset & \href{https://zenodo.org/records/10597292}{Zenodo} \\
\hline
\end{tabular}
\end{table}
\raggedbottom
\vspace{-1em}

\subsection{Methodology}

The proposed system consists of a two-stage pipeline for analyzing panoramic radiographs. The pipeline enables third molar localization, classification, and pericoronitis detection, supplemented by an explainability component.

Input images are converted to grayscale and resized to 832×832 pixels. The ROIs output by the first model are resized to 224×224 using center cropping before being passed to the second model.

Initially, YOLOv8 was used to process panoramic radiographs and detect third molars by generating bounding boxes. The YOLOv8 model is trained using images annotated by bounding boxes. It also performs two classification tasks: identifying the anatomical position (UR, UL, LL, LR) and determining the angulation based on Winter’s classification. Augmentation methods like mosaic, flips, rotation have been incorporated for better generalization~\citep{ultralytics2025augmentation}.

These ROIs are then used by the second-stage classifier, a modified ResNet-50 model, which outputs a binary label indicating the presence or absence of pericoronitis. Augmentations like flips and rotation were incorporated, whereas brightness and contrast augmentations were avoided due to feature loss.

\begin{figure}[H]
\centering
\includegraphics[width=0.85\textwidth]{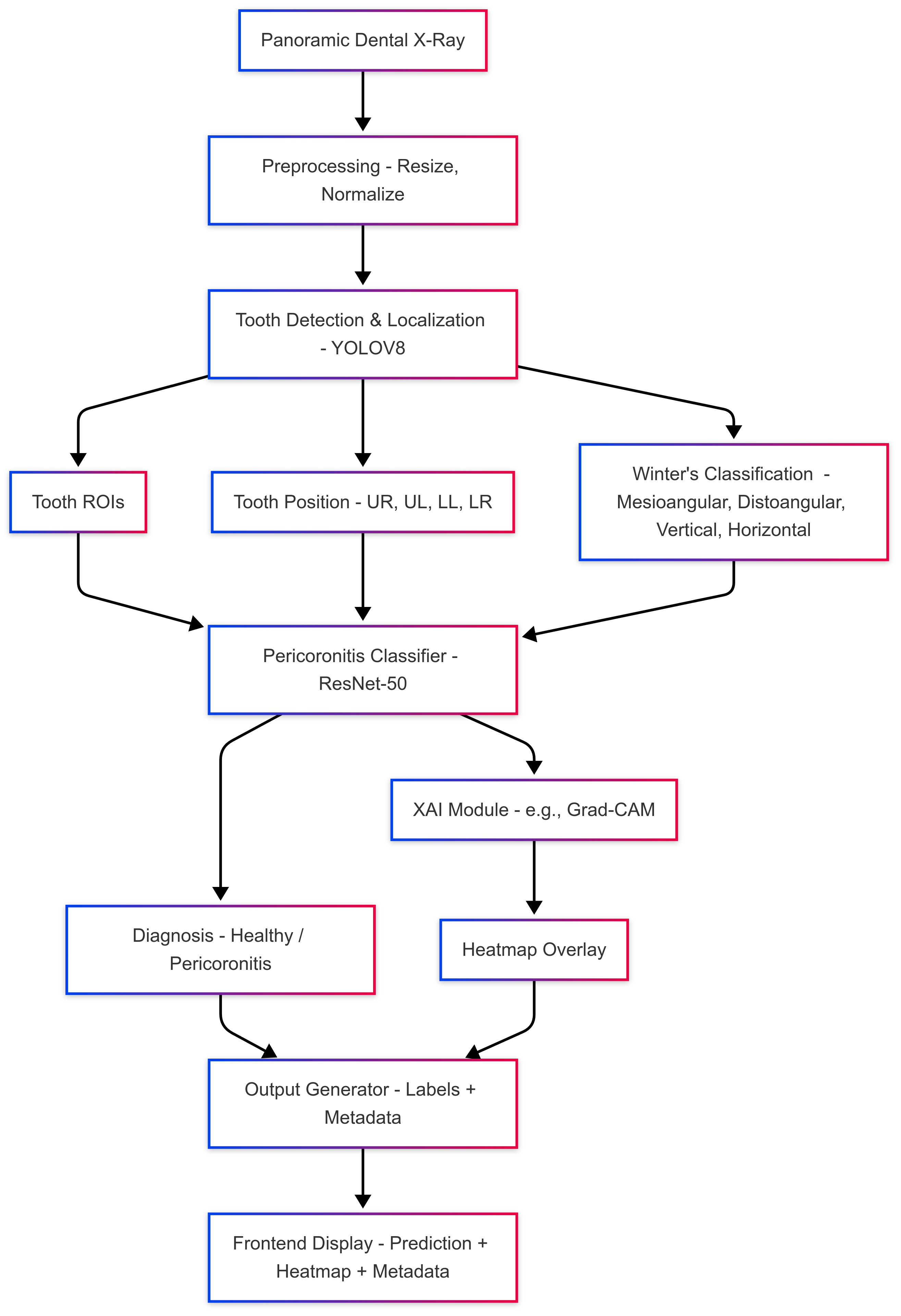}
\caption{End-to-end architecture of the proposed model pipeline showing (A) YOLOv8 localization, (B) ResNet-50 classification, and (C) Grad-CAM explainability.}
\label{fig:architecture}
\end{figure}

To improve transparency, Grad-CAM is applied to generate heatmaps over the ROIs using the last convolutional layer of the ResNet-50 model, visually highlighting influential regions in the model's decision process. An overview of the architecture is shown in Figure~\ref{fig:architecture}.

\section{Results}\label{sec4}

\begin{figure}[H]
\centering
\includegraphics[width=0.9\textwidth]{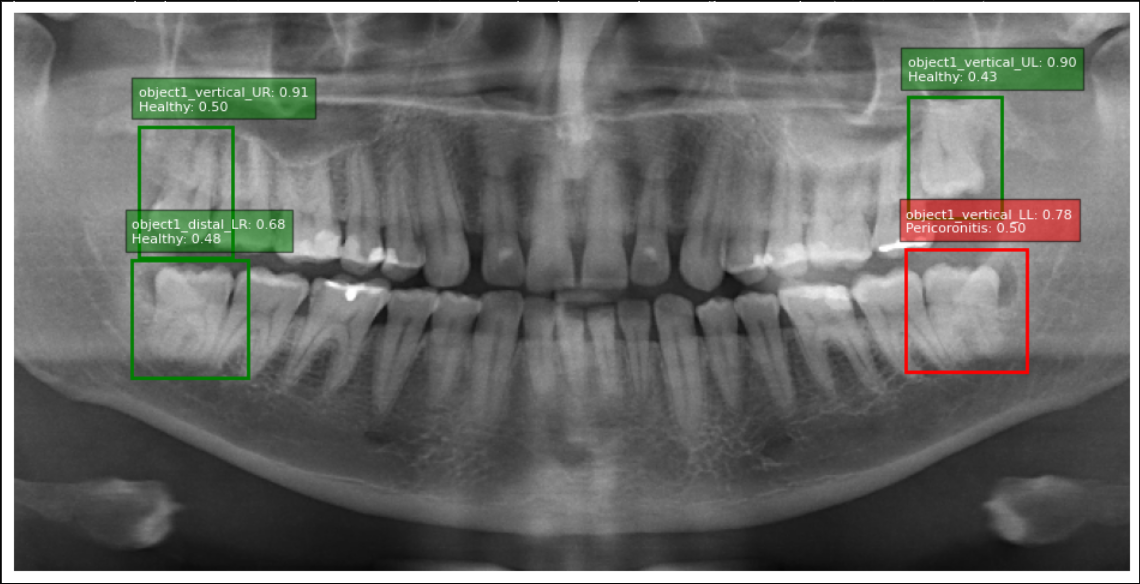}
\caption{Detection, anatomical localization and pathological classification of pericoronitis of third molars}
\label{fig:model}
\end{figure}

The proposed two-stage deep learning pipeline demonstrates strong potential for automating the detection and localization of pericoronitis in panoramic radiographs by integrating anatomical localization, pathological classification, and explainable AI.(Fig.~\ref{fig:model}).

\subsection{Technical Inference}

 The first stage, utilizing YOLOv8, achieved high detection precision (0.92) and mean average precision (mAP50 of 0.925)(Table~\ref{table:1}), indicating robust performance in accurately localizing third molars and extracting relevant anatomical features. The second-stage ResNet-50 classifier also performed well, with F1-scores of 0.88 for normal cases and 0.86 for pericoronitis (Table~\ref{table:2}), reflecting effective discrimination between healthy and affected teeth.

\vspace{-0.5em}  

\begin{table}[ht]
\caption{Detection performance of Model 1 (Localization Model) on the validation set.}\label{table:1}%
\begin{tabular}{@{}llllllll@{}}
\toprule
Model & Class & Images & Instances & Box(P) & R & mAP50 & mAP50-95 \\
\midrule
Model 1 & all & 155 & 444 & 0.92 & 0.70 & 0.925 & 0.806 \\
\bottomrule
\end{tabular}
\end{table}

\begin{table}[ht]
\caption{Classification performance of ResNet-based Model 2(Classification Model).}\label{table:2}%
\begin{tabular}{@{}llllll@{}}
\toprule
Model & Class & Precision & Recall & F1-score & Support \\
\midrule
Model 2 & Normal & 0.82 & 0.94 & 0.88 & 114 \\
~ & Pericoronitis & 0.93 & 0.81 & 0.86 & 119 \\
\bottomrule
\end{tabular}
\end{table}

\vspace{-0.5em}  

\begin{figure}[H]
\centering
\includegraphics[width=0.85\textwidth]{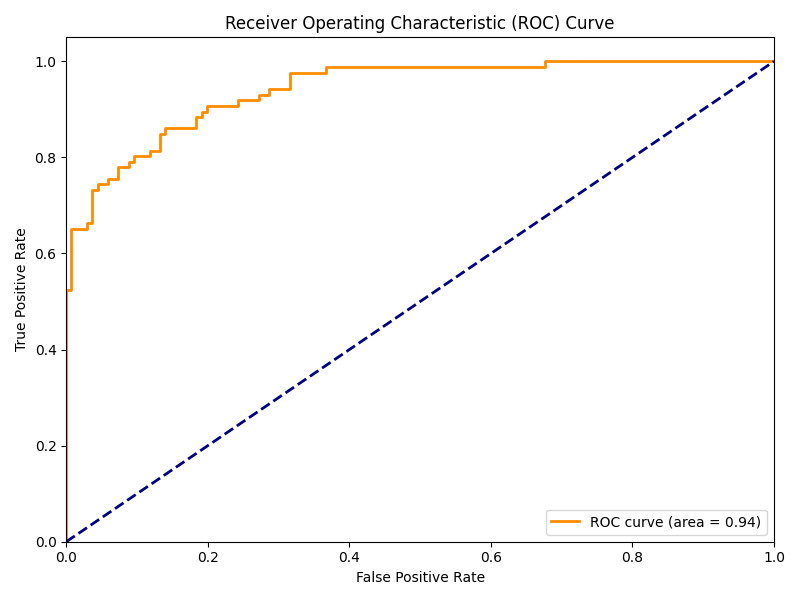}
\caption{ROC curve of the ResNet-50-based model on the test set, with AUC of 0.94 for classifying pericoronitis in panoramic radiographs.}
\label{fig:AUC}
\end{figure}

 The model achieved an AUC of 0.94 (Fig.~\ref{fig:AUC}). This result obtained is comparable to the current single model approach~\citep{vinayahalingam2021classification}

\begin{figure}[H]
\centering
\includegraphics[width=0.85\textwidth]{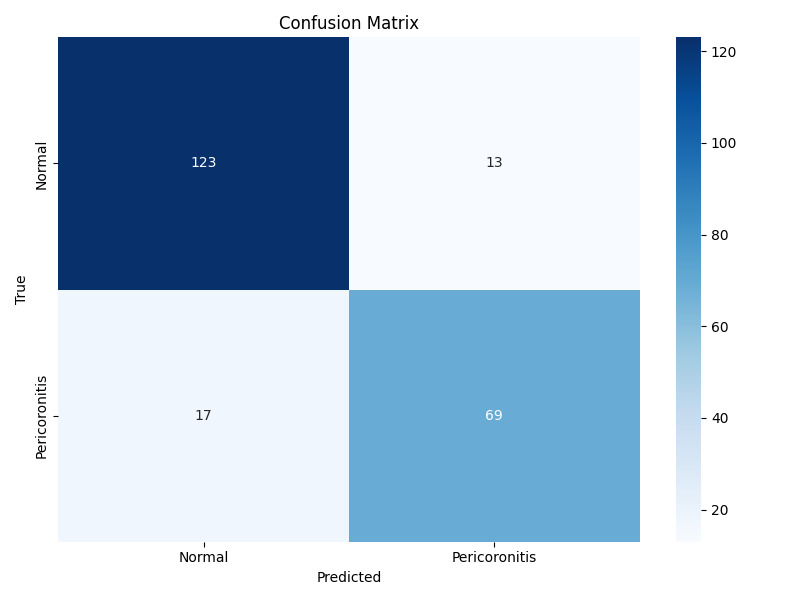}
\caption{Confusion matrix of Model 2: Classification Model}
\label{fig:matrix}
\end{figure}

Analysis of the 222 unseen samples yielded results comparable to those of the validation set, as illustrated in Fig.~\ref{fig:matrix}.

\subsection{Clinical Inference}

The dual-stage architecture replicates clinical reasoning by first identifying anatomical location of third molars and their Winter’s classification (vertical, mesioangular, horizontal, and distoangular), then analyzing radiological features of pericoronitis. The system efficiently distinguishes between healthy teeth and those affected by pericoronitis, which presents radiographically with distal/mesial bone loss, widened follicular space ($>2\text{mm}$), and pericoronal semilunar or circumferential radiolucent lesion. Panoramic radiographs reveal tooth angulation and marginal bone resorption Fig.~\ref{fig:model}, while advanced cases show abscess formation or osteomyelitis, requiring CT for early inflammation assessment.

\begin{figure}[H]
\centering
\begin{minipage}{0.9\textwidth}
  \centering
  \includegraphics[width=\textwidth]{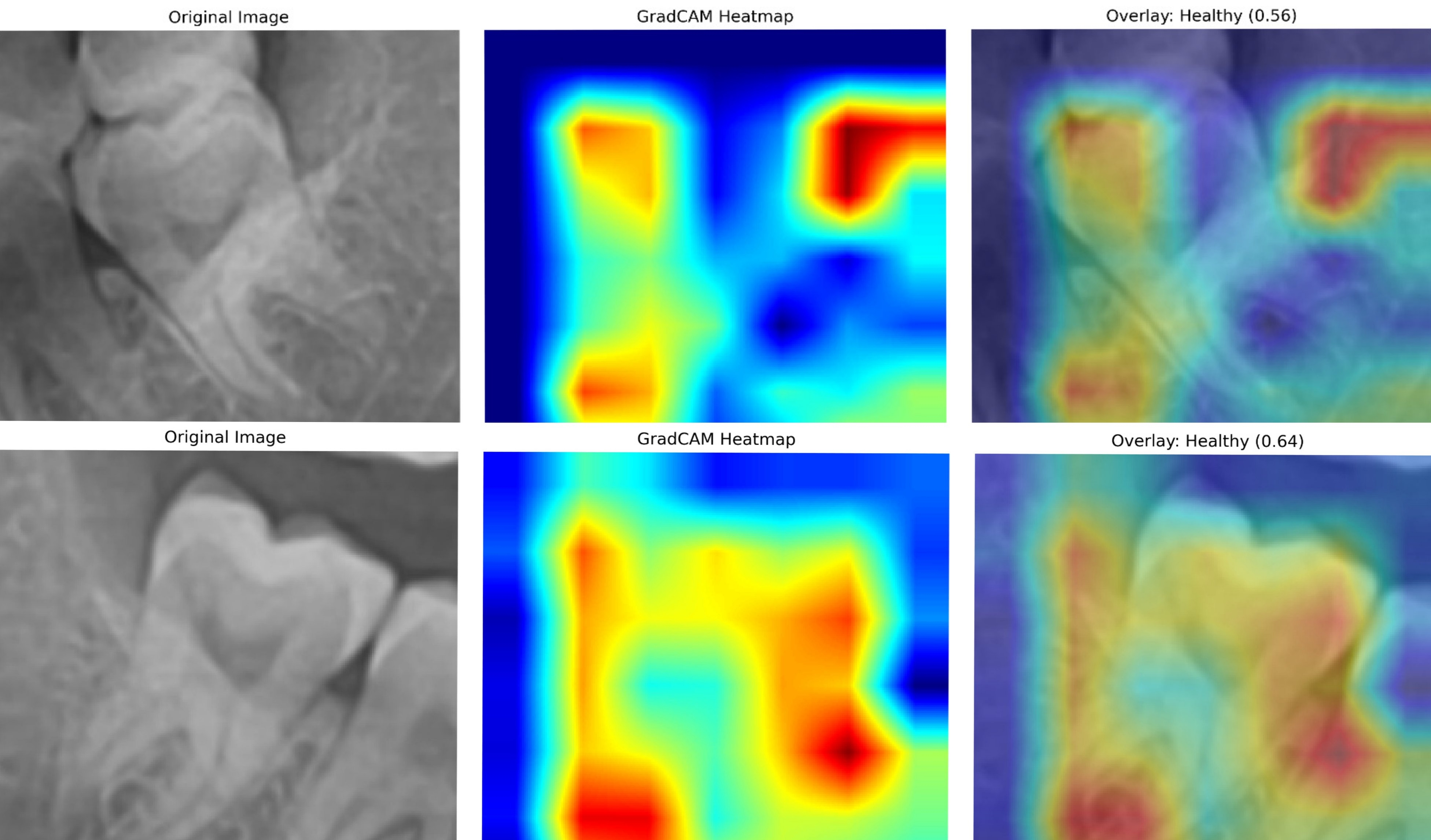}
  \caption{Grad-CAM visualizations of normal cases: (a) Mandibular left, (b) Mandibular right.}
  \label{fig:gradcam-normal}
\end{minipage}

\vspace{1em}

\begin{minipage}{0.9\textwidth}
  \centering
  \includegraphics[width=\textwidth]{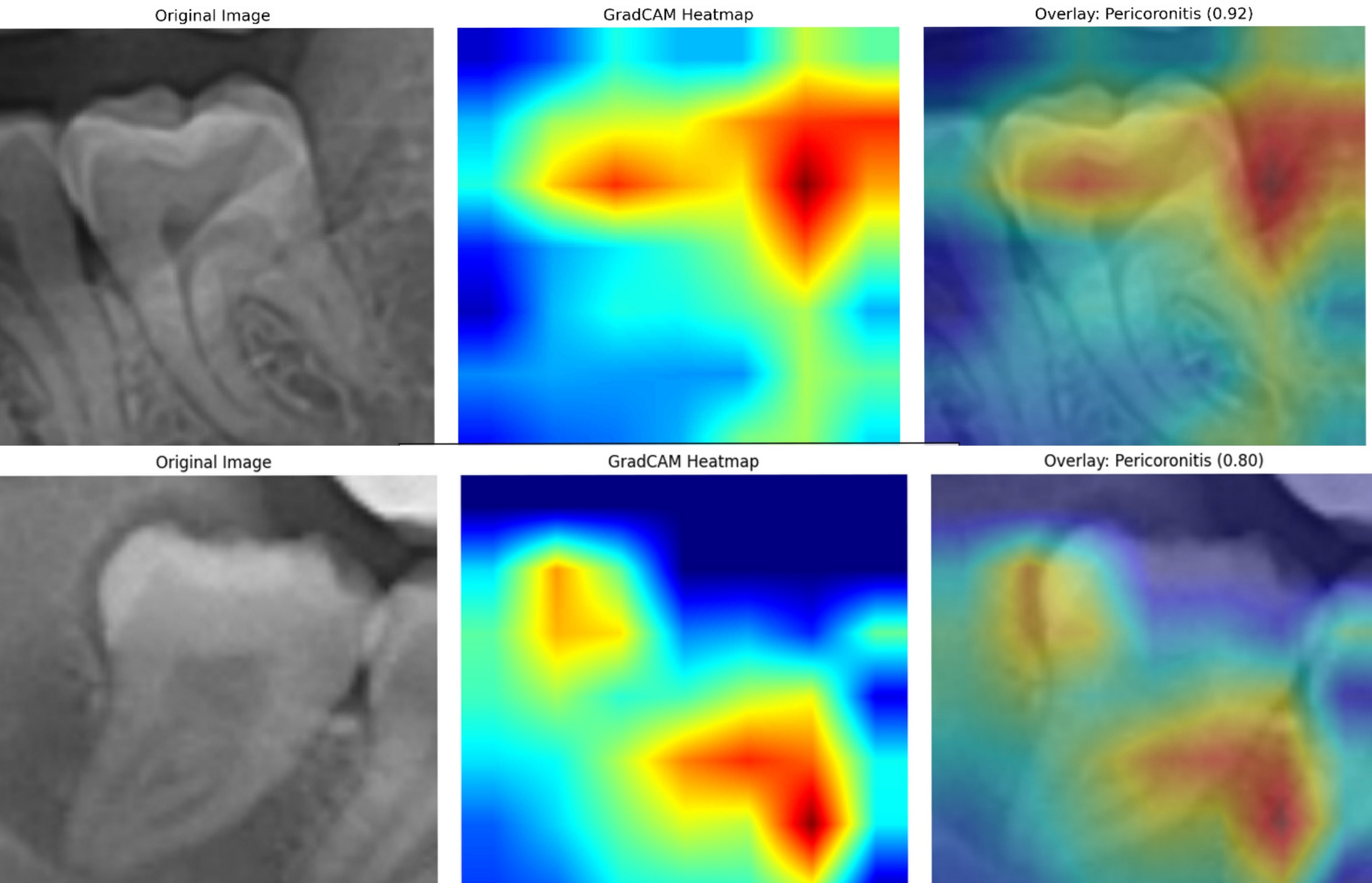}
  \caption{Grad-CAM visualizations of abnormal cases: (a) Mandibular left, (b) Mandibular right.}
  \label{fig:gradcam-abnormal}
\end{minipage}
\end{figure}

For normal teeth, Grad-CAM heatmaps highlighted healthy anatomical features, including the tooth structure, normal alveolar bone levels, and typical follicular space Fig.~\ref{fig:gradcam-normal}. The visualization indicates the model's attention to the absence of pathological indicators.

In radiologically radiographically suggestive of pericoronitis cases Fig.~\ref{fig:gradcam-abnormal}, overlaid Grad-CAM heatmaps intensely highlighted pericoronal radiolucent areas adjacent to the distal aspect of mandibular right . The hottest regions corresponded to key radiological features such as semilunar radiolucent aspect, widened follicular space and interdental osteolytic areas, suggesting adequate differentiation of osteolysis

Compared to existing models across tasks such as caries detection, lesion classification, segmentation, and periodontal staging, our pipeline demonstrates competitive performance (F1: 0.86–0.88) with integrated anatomical localization and explainability, highlighting its clinical relevance as depicted in Table~\ref{tab:ai_dental_dataset_models}.
\begin{sidewaystable}[htbp]
\caption{Comparison of the proposed explainable AI pipeline for pericoronitis detection with other published dental AI models across tasks}
\vspace{1em} 

\label{tab:ai_dental_dataset_models}
\begin{tabular*}{\textheight}{@{\extracolsep\fill} p{4.5cm} p{4.5cm} p{4.5cm} p{4.5cm}}
\toprule
\textbf{Dataset} & \textbf{Model} & \textbf{Key Metrics} & \textbf{Grad-CAM Usage} \\
\midrule
\textbf{Proposed Pipeline:}
1,190 OPGs (localization), 1,550 ROIs (classification), multi-center  & 
YOLOv8 (localization/classification), ResNet-50 (pericoronitis classification) & 
YOLOv8: 92\% precision, 92.5\% mAP50; ResNet-50: F1-score 0.88 (normal), 0.86 (pericoronitis); 84\% clinician agreement & 
Yes;reviewed by radiologists \\

500 PRs, 253 patients \cite{vinayahalingam2021classification} & 
MobileNetV2 & 
89\% accuracy (binary caries classification) & 
Yes \\

716 OPGs \cite{gwak2024attention}& 
ResNet50 + attention loss & 
91\% AUC (jaw lesion classification) & 
Yes \\

10,872 OPGs \cite{yu2022deep} & 
Two-branch CNN (classification + segmentation, MoCoV2 encoder) & 
83\% F1-score (general lesion detection) & 
Yes \\

13,870 images, 562 subjects \cite{oztekin2023explainable} & 
ResNet-50 & 
85\% accuracy (tooth-level caries) & 
Yes \\

400 OPGs, 780 lesions \cite{boztuna2024segmentation} & 
U²-Net & 
0.78 DSC (periapical lesion segmentation) & 
No \\

2,000 OPGs \cite{jundaeng2025artificial} & 
YOLOv8 & 
89\% mAP50 (periodontitis staging); κ=0.61 expert agreement & 
No \\

458 OPGs,\cite{lee2021deep} & 
DenseNet121 & 
87\% accuracy (SBC vs cyst/tumor) & 
No \\

14,438 clinical reports \cite{zhang2023grad} & 
ResNet18 (text) & 
Not image-based; text classification & 
Yes \\
\bottomrule
\end{tabular*}
\end{sidewaystable}

\section{Discussion}

This pipeline unifies both tooth localization and disease classification tasks, more closely mirroring the clinical workflow than prior studies that often focus on these in isolation (Table~\ref{tab:ai_dental_dataset_models}). By explicitly incorporating anatomical features, including quadrant and angulation, the model offers a richer diagnostic context, potentially reducing false positives and enhancing clinical relevance. The observed performance metrics demonstrate a favorable comparison with existing dental AI research

The integration of Grad-CAM-based explainability is a significant advancement. Visual heatmaps generated by Grad-CAM provide radiologists with valuable visual insights into the model’s decision-making process, fostering trust and confidence in the AI-based system~\citep{Elazab2024}. This feature not only supports clinical validation of AI outputs but also facilitates training and consensus-building among practitioners, especially in settings where dentomaxillofacial radiologists are promptly unavailable.

A qualitative assessment was conducted in collaboration with two experienced oral radiologists who reviewed a subset of 50 Grad-CAM overlays. Feedback indicated that in over 84\% of cases, the highlighted regions corresponded well with clinically relevant findings, thus supporting the utility of explainable AI in augmenting radiographic interpretation. Moreover, the visualization improved user trust in model outputs and was particularly beneficial in borderline or ambiguous cases. This integration of interpretability not only aligns with ethical AI deployment but also serves as an educational aid in training environments, making the pipeline suitable for broader clinical translation and adoption.

Despite promising results, several limitations should be acknowledged. First, the dataset size, while substantial, may not capture the full spectrum of anatomical and pathological variations present in broader populations. Second, the avoidance of brightness and contrast augmentations—while intended to preserve subtle radiographic features—may limit the model’s robustness to variations in acquisition quality across different imaging devices or settings. Most importantly, a limitation of this study is the absence of clinical symptom data; thus, the model is restricted to identifying radiographic correlates of pericoronitis and cannot distinguish between acute and chronic phases.

The proposed approach offers several avenues for future enhancement. Improving the quality of input panoramic radiographs—through higher-resolution imaging and standardized acquisition protocols—can enhance feature extraction, thereby enabling the application of brightness and contrast augmentations without significant risk of feature degradation. Additionally, the current binary classification (healthy vs. pericoronitis) can be extended into a fine-grained, multi-class diagnostic framework that differentiates varying severities of inflammation (e.g., mild, moderate, severe), enabling more nuanced and clinically actionable insights. Furthermore, for practical deployment in resource-constrained clinical environments, the model can be optimized for real-time inference on edge devices using techniques such as quantization and model pruning, ensuring faster and more efficient diagnostics without sacrificing accuracy.

\section{Conclusion}\label{sec5}

A two-model deep learning framework is introduced for localisation and detection of pericoronitis in partially erupted third molars using panoramic  radiographs, integrating object detection, classification, and explainability. The YOLOv8-based model effectively localizes third molars and classifies their anatomical features, while the ResNet-50-based classifier identifies the presence of pericoronitis.

The integration of explainable AI, particularly Grad-CAM, enhances clinical transparency by visualizing the features influencing predictions. This capability aids clinicians in understanding and trusting automated decisions. A blinded reader study with two experienced oral radiologists confirmed the clinical relevance of Grad-CAM overlays, showing 84\% alignment with expert interpretation and reinforced the model's applicability in real-world diagnostic workflows, especially in ambiguous cases.

Results indicate improved diagnostic accuracy and robustness across diverse datasets, suggesting the viability of this method in real-world settings. However, practical deployment faces challenges including YOLOv8’s hardware demands, regulatory approval hurdles, and the need to integrate clinical symptoms for comprehensive, real-world diagnostic accuracy. Future work will explore dataset expansion, inclusion of clinical metadata, and improvements in explainability mechanisms to support broader clinical adoption.These enhancements aim to improve generalizability. They will also enable real-time deployment on resource-constrained hardware, and support broader clinical translation of AI-assisted radiographic diagnosis.

\section{Declarations}\label{sec7}

\subsection{Funding} The authors declare that no funds, grants, or other support were received during the preparation of this manuscript.

\vspace{1em} 

\subsection{Competing Interest} The authors declare that they have no competing interests.

\vspace{1em} 
\subsection{Ethical Statement} This study did not involve human participants, animals, or clinical interventions, and thus ethical approval was not required.


\subsection{Informed Consent} This study did not involve human participants or identifiable data, and therefore informed consent was not required.




\subsection{Clinical Trial Registration} Not applicable – retrospective radiographic study.

\subsection{Data Availability Statement} 
The datasets curated for this study are available from the corresponding authors upon request. Also,
\begin{itemize}
    \item Dental OPG X-ray Dataset~\citep{imtkaggleteam_dental_opg_xray_2024} is publicly available at \url{https://www.kaggle.com/datasets/imtkaggleteam/dental-opg-xray-dataset}.
    \item Panoramic Dental X-ray dataset~\citep{amitabhabu_panoramic_dental_xray_2024} publicly available at \url{https://www.kaggle.com/datasets/amitabhabu/panoramic-dental-x-ray-images}
    \item Dental X-ray Images dataset~\citep{kayadibi_mtm_dataset_2024} publicly available at \url{https://data.mendeley.com/datasets/xn5bz6fdm6/1}
    \item STS-Tooth: A multi-modal dental dataset~\citep{wang2024sts_tooth} publicly available at \url{https://zenodo.org/records/10597292}
    
\end{itemize}

\vspace{1em} 

\bibliography{sn-bibliography}

\end{document}